\documentclass[conference,a4paper]{IEEEtran}
\IEEEoverridecommandlockouts
\usepackage{cite}
\usepackage{amsmath,amssymb,amsfonts}
\usepackage{algorithmic}
\usepackage{graphicx}
\usepackage{textcomp}
\usepackage{xcolor}

\usepackage{multirow}

\def\BibTeX{{\rm B\kern-.05em{\sc i\kern-.025em b}\kern-.08em
    T\kern-.1667em\lower.7ex\hbox{E}\kern-.125emX}}
\begin{document}

\title{Leveraging Temporal Joint Depths for \\Improving 3D Human Pose Estimation in Video\\
}

% \author{\IEEEauthorblockN{1\textsuperscript{st} Naoki Kato}
\author{\IEEEauthorblockN{Naoki Kato}
\IEEEauthorblockA{%\textit{dept. name of organization (of Aff.)} \\
\textit{Mobility Technologies Co., Ltd.}\\
% Tokyo, Japan \\
naoki.kato@mo-t.com
}
\and
% \IEEEauthorblockN{2\textsuperscript{nd} Hiroto Honda}
\IEEEauthorblockN{Hiroto Honda}
% \vspace{2mm}
\IEEEauthorblockA{%\textit{dept. name of organization (of Aff.)} \\
\textit{Mobility Technologies Co., Ltd.}\\
% Tokyo, Japan \\
hiroto.honda@mo-t.com
}
\and
% \IEEEauthorblockN{3\textsuperscript{rd} Yusuke Uchida}
\IEEEauthorblockN{Yusuke Uchida}
\IEEEauthorblockA{%\textit{dept. name of organization (of Aff.)} \\
\textit{Mobility Technologies Co., Ltd.}\\
% Tokyo, Japan \\
yusuke.uchida@mo-t.com
}
}

% \vspace{-40mm}

\maketitle

\setlength\textfloatsep{3mm}

% Variables
\newif\ifshowfig
\newif\ifshowtab
\newif\ifjapanese
\showfigtrue % Comment out this line to compile without figures
\showtabtrue % Comment out this line to compile without tables
% \japanesetrue % Comment out this line to compile in English

% -------------------------------------------------------------
\begin{abstract}
% -------------------------------------------------------------
\ifjapanese
3次元人物姿勢推定において、動画の各フレームで推定された2次元姿勢から3次元姿勢を推定するアプローチの有効性が確認されている。しかし、人物の外観情報を持たない2次元姿勢は関節点の奥行きに関して大きな曖昧性を持っている。本稿では、動画の各フレームで3次元姿勢を推定し、それらを前後フレームの推定結果を用いて補正するアプローチを提案する。これにより関節点の奥行きの曖昧性が軽減され、認識精度を改善することができる。
\else
The effectiveness of the approaches to predict 3D poses from 2D poses estimated in each frame of a video has been demonstrated for 3D human pose estimation. However, 2D poses without appearance information of persons have much ambiguity with respect to the joint depths. In this paper, we propose to estimate a 3D pose in each frame of a video and refine it considering temporal information. The proposed approach reduces the ambiguity of the joint depths and improves the 3D pose estimation accuracy.
\fi
\end{abstract}

% -------------------------------------------------------------

% -------------------------------------------------------------
\section{Introduction}
% -------------------------------------------------------------
% タスクの概要
\ifjapanese
3次元人物姿勢推定は単一または複数視点の画像や動画から人物関節点の3次元座標を特定するタスクであり、AR/VR、人物行動解析、車載カメラ映像解析などの応用先が存在する。本タスクの中でも単眼カメラ画像を入力とした3次元姿勢推定が特に盛んに研究されている。この場合、カメラから人物への奥行きおよび人物の大きさが不定となるため、基本的にはカメラ座標系において人物の腰を原点としたときの各関節点の相対座標を推定する問題設定となる。
\else
3D human pose estimation aims to localize human joints in a 3D coordinate system. It is important for many applications including AR/VR, human action analysis, and in-vehicle camera video analysis.
% 3D pose estimation from a monocular image is particularly actively studied. In this case, since the absolute depth and size of the person are uncertain, it is common to estimate the relative joint coordinates from a root joint (typically hip) in the camera coordinate system \cite{sun2018integral,tekin2016structured,pavlakos2017coarse}.
\fi
% タスクの課題：奥行きの不定性、データセットのバイアス
\ifjapanese
このように画像からの関節点の奥行きの推定には不定性が存在することが、タスクの難点の一つである。
また、関節点の3次元座標のアノテーションには一般的に実験室環境で構築されたモーションキャプチャシステムが必要となるため、大規模なデータセットの構築が困難である。背景や人物の服装や姿勢などの外観の多様性が低いデータセットを用いると、学習されたモデルの汎化性能を損なう恐れがあるという課題がある。
\else
One of the difficulties of this task is that there is ambiguity in the estimation of the depth of the joints from an image.
% In addition, since the annotation of the 3D coordinates of the joints generally requires a motion capture system constructed in a laboratory environment, building a large scale dataset is difficult. Training a model on such a dataset with low diversity in appearance, such as background, human clothing and pose may result in  poor generalization performance.
\fi
\ifjapanese
% 先行研究
近年の深層学習の研究の発展に伴い、3次元姿勢推定の精度も大きく向上している。
その中でも、タスクを画像からの2次元姿勢推定と、推定された2次元姿勢からの3次元姿勢推定に分割するアプローチの有効性が確認されている\cite{martinez2017simple}。2次元姿勢推定には多様性の高いデータセットを用いて学習された既製の2次元姿勢推定器を利用できる点、また3次元姿勢推定の際に画像の外観情報を敢えて使用しないことで過学習を軽減している点が本アプローチの有効性に寄与していると考えられる。
また、近年では動画における時系列情報の活用を図る研究が増えつつある。
% Pavlloらは動画の各フレームで推定された2次元姿勢を1次元畳み込みニューラルネットワークに入力することによる高精度な3次元姿勢推定手法を提案した\cite{pavllo20193d}。
それらの研究では、フレーム毎に推定された2次元姿勢をモデルの入力とするアプローチが主流となっている\cite{rayat2018exploiting,pavllo20193d,cai2019exploiting}。
\else
% With the recent development of deep neural networks, the performance of 3D pose estimation has been greatly improved. In particular, the approach that decompose the task into 2D pose estimation from an image and 3D pose estimation from the estimated 2D pose has proven to be effective \cite{martinez2017simple}.
% The availability of the off-the-shelf 2D pose estimator trained with highly diverse dataset, and avoiding overfitting by not using appearance information for estimating 3D pose contribute to the effectiveness of the approach.
In recent years, there has been an increasing number of studies on the use of temporal information in video. The most common approach in those studies is to utilize 2D poses estimated in each frame to predict final 3D pose \cite{rayat2018exploiting,pavllo20193d,cai2019exploiting}.
\fi

\ifjapanese
% 先行研究の課題
しかし、2次元姿勢は画像の外観情報を廃しているため、関節点の奥行きに関して大きな不定性を持っている。Pavlloらの実験において、正解の2次元姿勢を用いて推論を行っても完璧とは程遠い認識精度となったこと\cite{pavllo20193d}からも、2次元姿勢から3次元姿勢を推定するアプローチには精度的限界があると考えられる。
\else
However, since 2D pose does not include the appearance information, there is large ambiguity with respect to the depths of the joints. The experiments by Pavllo et al. \cite{pavllo20193d} suggest that estimating the 3D pose from the 2D pose is likely to have limited accuracy, even if the ground-truth 2D poses are used.
\fi

\ifjapanese
% 提案手法
% 概要
本稿では動画を入力した3次元姿勢推定において、動画の各フレームに対して3次元姿勢を推定し、それら複数フレームの推定結果を用いて最終的な3次元姿勢を推定するアプローチを提案する。
% 画像からの方が精度良く推定可能な3次元姿勢を入力にする
本手法は2次元姿勢よりも情報量の多い画像からの方が精度良く関節点の奥行きを推定できるという仮定に基づいている。2次元姿勢を中間的に使用する場合とは異なり、関節点の奥行きに関する情報を途中で損なうことのない推論が可能である。
% モデルについて（自由）
本アプローチにおいて、前段と後段それぞれのモデルには任意のものを使用することができる。我々は、前段モデルには2次元、3次元データセットを併用した学習が可能なIntegral Regression \cite{sun2018integral}を採用し、後段モデルには2次元姿勢からの3次元姿勢推定で有効性が確認されている1次元畳み込みニューラルネットワーク\cite{pavllo20193d}を使用する。
% 実験（TODO:もうちょい肉付け. 気付きとか）
公開データセットを用いた評価実験により提案手法の有効性を示す。
\else
In this paper, we propose a novel 3D pose estimation approach that first estimates a 3D pose for each frame (first stage) and then aggregates the multi-frame predictions for estimating the final 3D poses (second stage). This method is based on the assumption that the joint depths can be estimated more correctly from an image, which has more information than 2D pose. Unlike using 2D poses intermediately, it is possible to perform inferences that do not lose the joint depth information.
% In our approach, we can use arbitrary models for the first and second stage model respectively.
% We employ Integral Regression \cite{sun2018integral} for the first stage, which can be trained with both 2D and 3D datasets.
% We employ 1D ConvNet \cite{pavllo20193d}, which has been validated in 3D pose estimation from 2D poses, for the second stage.
We demonstrate the effectiveness of the proposed method through experiments on a public dataset.
\fi

% -----------------------------------------
% システム構成
% -----------------------------------------
\ifshowfig
\begin{figure}[t]
  \centering
  \includegraphics[width=0.80\linewidth]{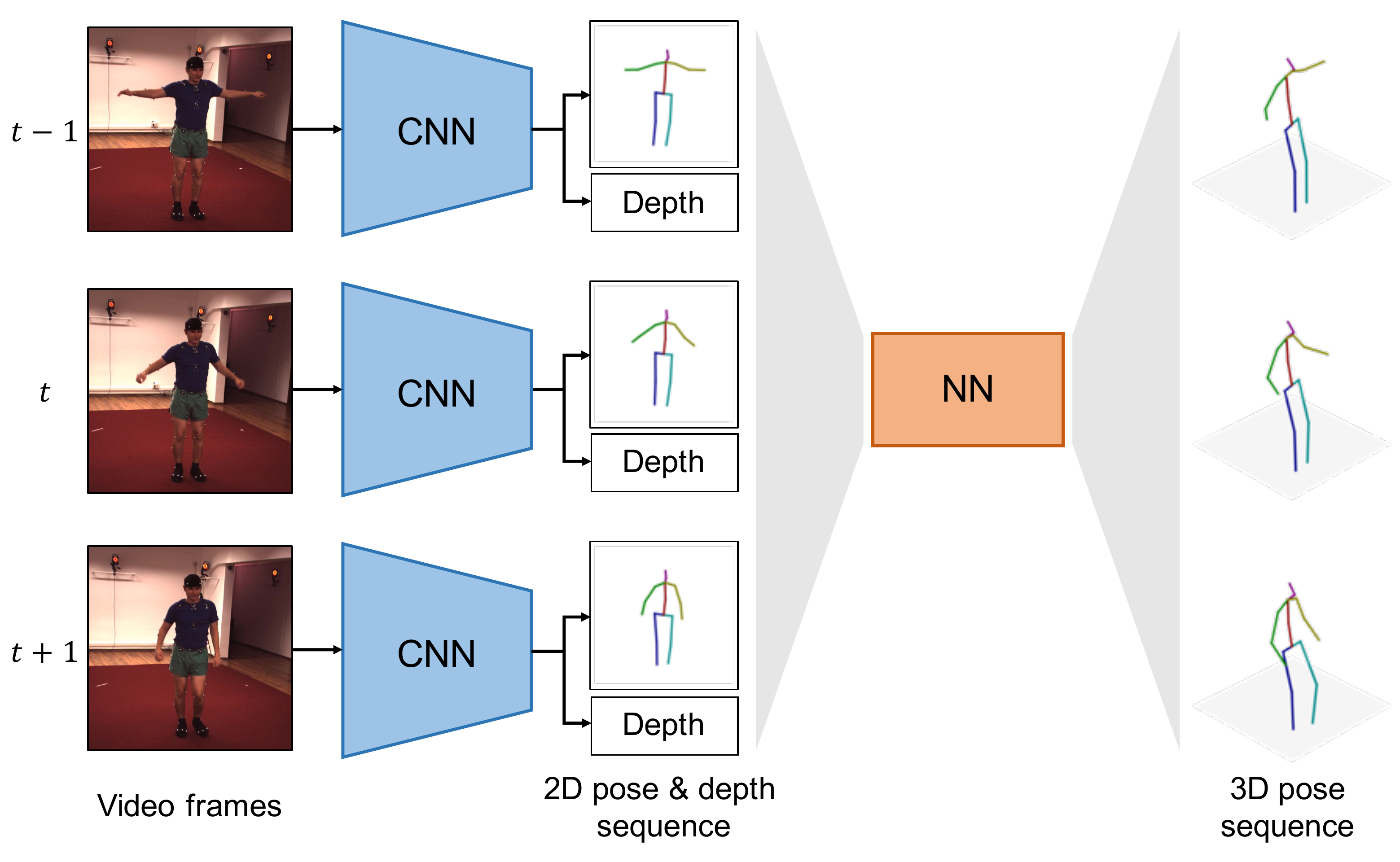}
  \caption{Overview of our approach.}
  \label{fig:approach}
\end{figure}
\fi

% -------------------------------------------------------------
\section{Proposed Method}
% -------------------------------------------------------------
\ifjapanese
% 手法の概要
図\ref{fig:approach}に提案手法の概要図を示す。
時刻$t$のフレームが高さ$H$、幅$W$の人物画像$I \in \mathbb{R}^{H \times W \times 3}$であるフレーム数$T$の人物動画$\{I_t\}$ ($t \in \{1,...,T\}$)が与えられたとき、入力動画から各フレームにおける人物関節点の3次元座標$y_t \in \mathbb{R}^{3J}$を推定することが我々の目的である。
ここで、$J$は推定対象の関節点数を表す。推定する関節点の3次元座標は、カメラ座標系において一つの関節点（主に腰が採用される）を基準とした各関節点の相対座標である\cite{tekin2016structured,martinez2017simple,pavllo20193d}。
\else
% Figure \ref{fig:approach} shows a schematic diagram of the proposed method.
% Given a video $\{I_t\}$ ($t \in \{1,...,T\}$) of length $T$ containing a human image $I_t \in \mathbb{R}^{H \times W \times 3}$ of size $H \times W$ at frame $t$, our objective is to estimate 3D coordinates of the human joints $y_t \in \mathbb{R}^{3J}$ for each frame where $J$ is the number of the joints.
% The 3D coordinates of the joints to estimate are defined as relative camera coordinate from the root joint (mainly hips are employed \cite{martinez2017simple,pavllo20193d}).

We show a schematic diagram of the proposed method in Figure \ref{fig:approach}.
Given a video $\{I_t\}$ ($t \in \{1,...,T\}$) of length $T$ containing a human image $I_t \in \mathbb{R}^{H \times W \times 3}$ of size $H \times W$ at frame $t$, our objective is to estimate 3D coordinates of the human joints $y_t \in \mathbb{R}^{3J}$ for each frame where $J$ is the number of the joints.
The target 3D coordinates of the joints are defined as relative coordinates from a root joint (usually hip is employed \cite{martinez2017simple,pavllo20193d}).
\fi

% -------------------------------------------------------------
% \subsection{Approach}
% -------------------------------------------------------------
\ifjapanese
% 既存手法
動画からの3次元姿勢推定の既存のアプローチとして、各フレームで推定された2次元姿勢のシーケンスをモデルに入力することで3次元姿勢を推定するアプローチが存在する\cite{rayat2018exploiting,pavllo20193d,cai2019exploiting}。
% 既存研究の利点
このアプローチではタスクの分割により個々のタスクを簡素化できるという利点があり、時系列情報を活用することの有効性も確認されている。
% 既存研究の問題点（奥行きの不定性が高い）
しかし、このアプローチで人物姿勢の中間表現として使用される2次元姿勢は、各関節点の奥行きに関する情報を欠いているため、2次元姿勢からの3次元姿勢推定は関節点の奥行きに関する不定性が大きくなるという問題がある。

% 画像からの方が正確なデプスを推定できる。単一画像の条件設定では、画像からの方が有効
関節点の奥行きを推定する際、関節点の奥行きに関する情報を廃した2次元姿勢よりも、人物の外観情報を持った画像を用いた方が問題の不定性が軽減されるため、関節点の奥行きは画像から推定するのが自然であると言える。
\else
% % 既存手法
% One of the existing approaches for estimating 3D poses from videos is to estimate the 3D poses from the sequence of 2D poses estimated at each frame \cite{rayat2018exploiting,pavllo20193d,cai2019exploiting}.
% % 既存研究の利点
% This simplifies individual tasks and the effectiveness of using temporal information has also been demonstrated.
% % 既存研究の問題点（奥行きの不定性が高い）
% However, since the 2D poses used as an intermediate representation of human pose in this approach lacks information about the depth of each joint, 3D pose estimation from 2D pose has the problem of large indeterminacy about the joint depth.
% It is natural choice to estimate the joint depths from images, since the problem is less ambiguous when using images with information about the appearance of the person rather than 2D poses that lacks information about joint depths.

One of the existing approaches for estimating 3D poses in video is to estimate the 3D poses from the 2D pose sequences estimated at each frame \cite{rayat2018exploiting,pavllo20193d,cai2019exploiting}. This simplifies the task and the effectiveness of exploiting temporal information has been demonstrated. However, since 2D poses used as intermediate representations of human poses in this approach lack the depth information of the joints, 3D pose estimation from 2D pose has the large ambiguity of joint depths. To resolve the issue, it is a natural choice to estimate the joint depths from images which have rich person appearance information.
\fi

\ifjapanese
% 提案手法
上記考察を踏まえ、我々は上述の2段階の3次元姿勢推定の枠組みにおいて、関節点のデプスを中間情報として用いることを提案する。
各段階のモデルをそれぞれ前段モデル、後段モデルと呼称する。
前段モデルは時刻$t$のフレーム$I_t$を入力に各関節点の画像座標およびデプスから成る3次元姿勢$x_t \in \mathbb{R}^{3J}$を推定し、
後段モデルは前段モデルで推定された3次元姿勢のシーケンス$\{x_t\}$ ($t \in \{1,...,T\}$)を入力に最終的な3次元姿勢$y_t$を推定する。
% 提案手法の利点
本手法は3次元姿勢を時系列情報を用いて補正するアプローチであると見做すこともできる。
2次元姿勢のみを中間情報として用いる手法とは異なり、関節点の奥行きに関する情報を推論の途中で損なわないという利点がある。
本アプローチでは、前段モデル、後段モデル共に任意のモデルを使用することができる。
次節以降では、前段モデルと後段モデルとして使用するそれぞれのモデルの詳細を述べる。
\else
% Based on the above considerations, we propose to use the depths of the joints as intermediate information in the framework of the above two-stage approach.
% The first stage predicts 3D pose $x_t \in \mathbb{R}^{3J}$ composed of image coordinates and depths of joints from each frame $I_t$, then the second stage estimates the final 3D pose $y_t$ from the sequence of 3D poses $\{x_t\}$ ($t \in \{1,...,T\}$) estimated by the first stage.
% % 提案手法の利点
% This approach can be regarded as an approach for refining the 3D pose using temporal information.
% The advantage of our method is that the information about the joint depths is not lost in the middle of the inference, unlike the approach which uses only 2D pose as intermediate representation. In this approach, any model can be used for both the first and second stage.

Based on the above considerations, we propose to use the depths of joints as an intermediate representation in the framework of the above two-stage method. The first stage predicts a 3D pose $x_t \in \mathbb{R}^{3J}$ comprised of image coordinates and depths of joints in each frame $I_t$, then the second stage predicts the final 3D pose $y_t$ from the 3D pose sequences $\{x_t\}$ ($t \in \{1,...,T\}$) estimated by the first stage.
This approach can be regarded as refining 3D poses using temporal information.
The advantage of the method is that the joint depth information is not lost in the middle of the inference, unlike the approach which uses only 2D poses as an intermediate representation.
% In this approach, we can use arbitrary models for both stages.
\fi

\ifjapanese
\noindent {\bf First stage}\,
% 概要
動画の各フレームから3次元姿勢を推定する前段モデルには、Integral Regression \cite{sun2018integral}を採用する。このモデルは画像を入力に関節点毎にサイズ$w \times h \times d$の3次元ヒートマップを出力する。ここで、ヒートマップの$x$-$y$方向のグリッドは画像の$x$-$y$方向をそれぞれ$w, h$段階に一様に離散化し、$z$方向のグリッドは腰を基準とした $[-D/2, D/2]$ mmの奥行きを$d$段階に一様に離散化したものとなっている。最終的な関節点の3次元座標の推定結果はヒートマップの重心座標から算出される。このモデルは2次元、3次元データセットを併用した学習が可能であり、外観の多様性に富む2次元データセットの併用により、3次元データセットで問題となる過学習を軽減することができる。
% GTの腰の奥行きを使うと最終的な3次元姿勢が求められる
% 損失関数の話

% モデル構造の詳細
我々はモデルのバックボーンにResNet-50 \cite{he2016deep}を使用し、カーネルサイズ4の転置畳み込み層を3層用いて関節点毎に$z$方向のチャネル数$d=72$の3次元ヒートマップを出力する。また、ヒートマップの$z$方向の大きさに対応する空間の奥行きを$D=1,500$ mmとする。
\else
% % 概要
% We employ Integral Regression \cite{sun2018integral} as the first stage that predicts 3D pose in each frame.
% Given a human image, this model outputs a 3D heatmap of size $w \times h \times d$ for each joint.
% Here, the $x$-$y$ grid uniformly discretize $x$-$y$ coordinates of the image and $z$ grid uniformly discretize $[-D/2, D/ 2]$ mm centered at the root joint.
% The final estimates of the joint coordinates are calculated from the centroid of the heatmap.
% The model can be trained using both 2D and 3D datasets. Using 2D datasets, which is diverse in apprearance, reduce overfitting problematic on the training with 3D datasets.
%
% % モデル構造の詳細
% We use ResNet-50 \cite{he2016deep} as the backbone and apply three transposed convolution layers of kernel size 4 to output 3D heatmaps where the number of channels in the $z$ axis is $d=72$.
% We set $D=1500$ mm.
% % We set the depth of the space corresponding to the size of the heatmap in the $z$ direction $D=1,500$ mm.
\noindent {\bf First stage.}\,
We employ Integral Regression \cite{sun2018integral} for the first stage, which predicts 3D pose in each frame. Given a human image, this model outputs a 3D heatmap of size $w \times h \times d$ for each joint. Here, the $x$-$y$ grid uniformly discretize the image coordinates and the $z$ grid uniformly discretize $[-D/2, D/ 2]$ mm centered at the root joint. The final estimates of the joint coordinates are computed from the centroids of the heatmaps. The model can be trained using both 2D and 3D datasets. By utilizing a 2D dataset which is diverse in appearance, we can mitigate overfitting which is problematic when training on 3D datasets.
We use ResNet-50 \cite{he2016deep} as the backbone and apply three deconvolution layers of kernel size 4 to output 3D heatmaps where the number of channels for the $z$ axis is $d=72$. We set $D=1500$ mm.
\fi

\ifjapanese
\noindent {\bf Second stage}\,
% 概要
3次元姿勢のシーケンスから最終的な3次元姿勢を推定する後段モデルには、2次元姿勢のシーケンスを入力とする手法\cite{pavllo20193d}で有効性が確認されている1次元畳み込みニューラルネットワークを採用する。
% TODO: 入出力（2D poseは[-1, 1]に正規化、デプスはメートル）
% モデル構造の詳細
モデルの入力層はカーネルサイズ$W$、出力チャネル数$C$の時系列方向への畳み込み層である。その後、カーネルサイズ$W$、dilation $D = W^B$の畳み込み層およびカーネルサイズ1の畳み込み層から成るresidual blockが$B$ブロック連なっている。出力層はカーネルサイズ1の畳み込み層であり、フレーム毎にチャネル数$3J$の推定姿勢が出力される。出力層を除く全ての畳み込み層の後にはbatch normalization、ReLU、dropout rate $p$のdropout層が連なっている。このモデルの受容野は$W^{B+1}$である。特に断りのない限り、我々は$W=3$、$B=4$、$C=1024$、$p=0.25$のモデルを使用する。
% TODO: 損失関数の話
\else
% We employ 1D ConvNet as the second stage, which is effective for exploiting sequences of 2D pose \cite{pavllo20193d}.
% The first layer of the model is a convolution layer in the temporal direction with kernel size $W$ and number of output channels $C$. After that, a residual block consisting of a convolutional layer of kernel size $W$, dilation $D = W^B$ and a convolutional layer of kernel size 1 is followed $B$ blocks. The output layer is a convolutional layer with a kernel size 1 and outputs estimated pose of $3J$ channels for each frame. All the convolutional layers except the output layer are followed by a batch normalization, ReLU and dropout layers with a dropout rate $p$. The receptive fields of this model is $W^{B+1}$. Unless otherwise noted, we set $W=3$, $B=4$, $C=1024$, and $p=0.25$.

\noindent {\bf Second stage.}\,
We employ 1D ConvNet as the second stage, which is effective for exploiting 2D pose sequences \cite{pavllo20193d}. The input layer is a temporal convolution with kernel size $W$ and output channels $C$. This is followed by $B$ residual blocks, each of which is composed of a 1D convolution with kernel size $W$, dilation $D = W^B$ and another convolution with kernel size 1. The output layer is a convolution with kernel size 1, which outputs an estimated pose with $3J$ channels for each frame. All the convolutional layers except the last layer are followed by batch normalization, ReLU and a dropout layer with a dropout rate $p$. The temporal receptive fields of this model is $W^{B+1}$. Unless otherwise noted, we set $W=3$, $B=4$, $C=1024$, and $p=0.25$.
\fi

\ifjapanese
% \noindent {\bf データ拡張}\,
% モデルを2段階に分割する場合、後段モデルにおけるtrainとtestの入力データの乖離が問題となる。データ拡張が重要
% 本手法のように2段階のモデルを使用する場合、
前段モデルの推定結果を用いて後段モデルを学習するが、このとき、
後段モデル学習時の入力は前段モデルの学習データに対する推定結果、テスト時の入力は前段モデルの未知データに対する推定結果となるため、学習時とテスト時の入力データの分布に乖離が生じ、モデルの性能が損なわれることが懸念される。そこで我々は後段モデル学習時、入力2次元姿勢およびデプスにガウシアンノイズを付加することで、未知データに対する前段モデルの推定結果を再現するデータ拡張を行い、モデルの性能向上を図る。
\else
We use the predictions of the first stage to train the second stage. In this case, the input of the second stage at training time is the predictions from the model which has been supervised by the training data for the first stage. However at test time, the input is the predictions on unseen data, so there is a concern that the distribution of the input data at training and test time may deviate and the performance of the model may be degraded. Therefore, we apply data augmentation that adds Gaussian noise to the input 2D poses and depths during training of the second stage, to reproduce the test-time predictions of the first stage to enhance model performance.
\fi

% -------------------------------------------------------------
\section{Experiments}
% -------------------------------------------------------------
% \subsection{Datasets and Evaluation Metrics}
\noindent {\bf Datasets and Evaluation Metrics.}\,
% -------------------------------------------------------------
\ifjapanese
% データセット
3次元姿勢データセットの中で最大級のデータセットである、Human3.6Mデータセット\cite{ionescu2013human3}を用いて評価実験を行う。このデータセットは実験室環境において11人の被験者を撮影した計約360万フレームの動画から成る。被験者の内7人物の関節点の3次元座標がアノテーションされており、各人物が15種類の行動を取る様子が4台の同期されたカメラで撮影されている。標準的な評価方法\cite{pavlakos2017coarse,martinez2017simple}に則り、5人物 (S1, S5, S6, S7, S8) を用いてモデルを学習し、2人物 (S9, S11) を用いてモデルを評価する。17関節点を使用し、全ての行動を用いて1つのモデルを学習する。
前段モデルとして使用するIntegral Regressionの学習にはHuman3.6Mデータセットに加え、2次元姿勢データセットであるCOCO 2017データセット\cite{lin2014microsoft}とMPII Human Poseデータセット\cite{andriluka20142d}を使用する。
\else
We conduct experiments using Human3.6M \cite{ionescu2013human3}.
% This dataset consists of 3.6 million images featuring 11 subjects in a laboratory environment.
% The 3D joint coordinates of 7 subjects are annotated.
% Each person is captured by 4 synchronized cameras as they perform 15 different activities.
Following the standard evaluation protocol \cite{martinez2017simple,sun2018integral,pavllo20193d}, we use 5 subjects (S1, S5, S6, S7, S8) for training and 2 subjects (S9, S11) for evaluation.
% We use 17 joints and all the actions to train a single model.
Along with Human3.6M, we use 2D datasets of COCO \cite{lin2014microsoft} and MPII \cite{andriluka20142d} to train Integral Regression.
\fi

\ifjapanese
% 評価指標・評価手順
Human3.6Mデータセットでの評価で一般的に用いられる2種類の評価プロトコルを用いてモデルの性能を評価する。プロトコル1は、関節点の推定座標と正解座標の距離を全ての関節点およびサンプルで平均することにより算出されるMean Per Joint Position Error (MPJPE)である。プロトコル2は、正解姿勢に合致するよう推定姿勢を並進移動、回転、スケーリングした後で算出される同様の評価指標であり、P-MPJPEとも呼ばれる。
\else
% We consider two evaluation protocols commonly used on the Human3.6M.
% Protocol \#1 is Mean Per Joint Position Error (MPJPE) calculated by averaging the distances between the predicted and ground-truth coordinates.
% Protocol \#2 is the error after alignment with the ground-truth in translation, rotation, and scale.

We consider two evaluation protocols commonly used for the evaluation on Human3.6M. Protocol 1 is Mean Per Joint Position Error (MPJPE) calculated by averaging the distances between the predicted and ground-truth coordinates. Protocol 2 is also MPJPE but after alignment with the ground-truth in translation, rotation, and scale.
\fi

\ifshowtab
\begin{table}
\centering
\caption{Impact on MPJPE by input and receptive field.}
\label{tab:evaluation_results}
% \scalebox{1.0}{
\begin{tabular}{l|cccc}
\multicolumn{1}{c}{\multirow{2}{*}{Input}}
  & \multicolumn{4}{|c}{Receptive field} \\ \cline{2-5}
                       & 1    & 27   & 81   & 243  \\ \hline \hline
2D pose                          & 49.5 & 48.8 & 47.7 & 47.6 \\
2D pose + depth                  & 50.4 & 48.5 & 48.4 & 47.8 \\
2D pose + depth ($\sigma = 0.1$) & 48.4 & 46.4 & 45.9 & 45.6 \\
GT 2D pose                       & 38.7 & 37.3 & -    & 36.3 \\
GT 2D pose + GT depth            & 20.2 & 13.4 & -    & 15.9 \\
% GT 3D pose                       & 17.7 & 10.1 & -    & 11.8 \\
\end{tabular}
% }
\end{table}
\fi

% -----------------------------------------
% 既存研究との性能比較
% -----------------------------------------
\ifshowtab
\begin{table}[t]
\caption{The evaluation results on Human3.6M dataset.}
% $(\dagger)$ uses temporal information.
\label{tab:comparison}
\centering
% {\scriptsize
% \begin{tabular}{lcc} %\bottomrule[0.3mm] % p{1.6em} \hfil
% Method & P1 & P2 \\ \hline
% Martinez et al. \cite{martinez2017simple}        & 62.9 & 47.7 \\
% Sun et al. \cite{sun2018integral}                & 49.6 & 40.6 \\
% % Lee et al. \cite{lee2018propagating} $(\dagger)$ & 52.8 & 43.4 \\
% Cai et al. \cite{cai2019exploiting} $(\dagger)$  & 48.8 & 39.0 \\
% Pavllo et al. \cite{pavllo20193d} $(\dagger)$    & 46.8 & 36.5 \\ \hline
% Ours $(\dagger)$                                 & {\bf 45.6} & {\bf 34.8} \\ %\bottomrule[0.3mm]
% \end{tabular}
\begin{tabular}{lcc} %\bottomrule[0.3mm] % p{1.6em} \hfil
Method & Protocol 1 & Protocol 2 \\ \hline \hline
Martinez et al. \cite{martinez2017simple}        & 62.9 & 47.7 \\
Sun et al. \cite{sun2018integral}                & 49.6 & 40.6 \\
% Lee et al. \cite{lee2018propagating} $(\dagger)$ & 52.8 & 43.4 \\
Cai et al. \cite{cai2019exploiting}  & 48.8 & 39.0 \\
Pavllo et al. \cite{pavllo20193d}    & 46.8 & 36.5 \\ \hline
Ours                                 & {\bf 45.6} & {\bf 34.8} \\ %\bottomrule[0.3mm]
\end{tabular}
% }
\end{table}
\fi

% -------------------------------------------------------------
% \subsection{Results}
\noindent {\bf Results.}\,
% -------------------------------------------------------------
\ifjapanese
\noindent {\bf 入力および受容野による性能変化}\,
各入力および受容野のモデルの評価結果を表\ref{tab:evaluation_results}に示す。表中の受容野が1、27、81、243のモデルは、それぞれ$(W, B)$が$(1, 2)$, $(3, 2)$, $(3, 3)$, $(3, 4)$のモデルである。
% デプスの入力
データ拡張を行わずにデプスを入力に付加した場合、2次元姿勢を入力とした場合と同程度のエラーとなった。これは学習時とテスト時のデプスの分布に乖離があり過学習が生じたためであると考えられる。一方、デプスにガウシアンノイズを加えて学習したモデルは全ての受容野において2次元姿勢を入力した場合よりも低いエラーとなった。受容野が243のとき、エラーは最も低い45.6 mmとなり、2次元姿勢を入力した場合と比べ2.0 mmエラーが削減された。この結果より、デプスを後段モデルの入力にすることの有効性が確認された。
% フレーム数を増やすことで性能が向上する
また、いずれの入力の場合も、受容野の大きなモデルほどエラーが低くなる傾向となった。
我々の学習したIntegral RegressionのMPJPEは48.9 mmであり、後段モデルの使用によりエラーが最大3.3 mm削減された。この結果は前後フレームの推定結果を活用することで単一画像3次元姿勢推定器の推定結果を適切に補正できることを示している。

% GTの結果に関する考察
% 正解2次元姿勢、正解2次元姿勢と正解デプスを入力した場合の最も低いエラーはそれぞれ37.3 mm、13.4 mmとなった。正解デプスを入力することでエラーが64\%削減されており、デプスの入力によりエラーの下界が大きく減少することが分かった。このことから、提案したアプローチには更なるエラー削減の余地があると言える。
正解2次元姿勢、また正解2次元姿勢および正解デプスを入力した場合の最も低いエラーはそれぞれ37.3 mm、13.4 mmとなり、入力の2次元姿勢が正確になることでモデルの性能が大きく改善することが確認された。また、正解デプスを入力することでエラーがさらに64\%削減されており、デプスの入力によりエラーの下界が大きく減少することが分かった。このことから、提案したアプローチには更なるエラー削減の余地があると言える。
\else
% \noindent {\bf Model input and receptive field}\,
The evaluation results for different input types and temporal receptive fields of the second-stage model are shown in Table \ref{tab:evaluation_results}. The models with receptive fields of 1, 27, 81, and 243 in the table have $(W, B)$ of $(1, 2)$, $(3, 2)$, $(3, 3)$ and $(3, 4)$ respectively. When joint depths are directly added to the inputs without data augmentation, the error is comparable to the result when only 2D poses are used as input. This may be due to overfitting caused by a discrepancy between the distribution of depths during training and test time. On the other hand, the model trained with Gaussian noise on depths resulted in lower errors in all receptive fields than 2D poses. This result demonstrates the effectiveness of using depths as inputs of the second stage with an appropriate data augmentation.

Our implementation of Integral Regression has MPJPE of 48.9 mm at the first stage, and the errors are reduced by up to 3.3 mm at the second stage, showing that the prediction results of a single-frame 3D pose estimator can be properly refined by utilizing temporal information.

The lowest error when using the ground-truth 2D pose as input is 36.3 mm, showing that the error of the system is greatly reduced by the accurate 2D pose. Adding ground-truth depths further reduce errors by 63\%, indicating that exploiting the depths significantly reduces the lower bound of errors.
\fi

\ifjapanese
\noindent {\bf 既存手法との性能比較}\,
提案手法と既存手法の性能比較結果を表\ref{tab:comparison}に示す。
両プロトコルにおいて、提案手法のエラーの平均が全ての比較手法を下回る結果となった。提案手法は比較手法の中で最もエラーの平均が低いPavlloらの手法と比べ、プロトコル1で1.2 mm、プロトコル2で1.7 mm低いエラーとなった。
\else
% \noindent {\bf Comparison with other works}\,
The evaluation results of our method and existing monocular methods are shown in Table \ref{tab:comparison}. For both protocols, our approach outperforms all the comparative methods.
% Compared to the method proposed by Pavllo et al.,
% % that has the lowest error among the comparative methods,
% our model achieves 1.2 mm lower error in protocol 1 and 1.7 mm lower error in protocol 2.
\fi

% -------------------------------------------------------------
\section{Conclusion}
% -------------------------------------------------------------
\ifjapanese
本稿では動画を入力とする2段階のモデルから成る3次元姿勢推定の枠組みにおいて、2次元姿勢に加えて関節点のデプスを人物姿勢の中間情報として用いるアプローチを提案した。評価実験では、未知データに対する前段モデルの推定結果を再現するデータ拡張を施すことで、デプスを後段モデルの入力に加えることにより認識性能が改善されることを確認した。
% 展望：遮蔽の扱い、後段モデルの構造、汎化性能
2段階のモデルを学習する際の適切なデータ拡張手法の確立、また関節点の遮蔽を明示的に扱うモデルの枠組みの提案が今後の展望として挙げられる。
\else
In this paper we proposed a two-stage 3D pose estimation pipeline in video that uses a joint depth sequence as an intermediate representation for the human pose in addition to a 2D pose sequence. In the evaluation experiments, we observe that adding depth to the input of the second stage reduces the 3D joint localization error, indicating that our pipeline appropriately refine 3D poses leveraging temporal information.
\fi

% -------------------------------------------------------------

\bibliographystyle{IEEEtran}
\bibliography{myref}

\end{document}